\documentclass[lettersize,journal]{IEEEtran}
\usepackage{amsmath,amsfonts}
\usepackage{algorithmic}
\usepackage{algorithm}
\usepackage{array}
\usepackage[caption=false,font=normalsize,labelfont=sf,textfont=sf]{subfig}
\usepackage{textcomp}
\usepackage{stfloats}
\usepackage{url}
\usepackage{verbatim}
\usepackage{graphicx}
\usepackage{cite}
\usepackage{mathtools}
\usepackage{mathrsfs}
\usepackage{float}
\usepackage{makecell}
\usepackage{booktabs}
\usepackage{multirow}
\usepackage{caption}
\usepackage[switch]{lineno}
\usepackage{color}

\begin{document}

\title{Spatial Temporal Attention based Target Vehicle\\ Trajectory Prediction for Internet of Vehicles}

\author{Ouhan Huang, Huanle Rao,~\IEEEmembership{Member,~IEEE}, Xiaowen Cai, Tianyun Wang, Aolong Sun, Sizhe Xing, Yifan Sun and Gangyong Jia,~\IEEEmembership{Member,~IEEE}
\thanks{Manuscript received April 19, 2021; revised August 16, 2021. The work was supported by the National Natural Science Foundation of China under Grant No.U20A20386, Zhejiang Key Research and Development Program under Grant No.2023C03194 and No.2020C01050, Zhejiang Natural Science Foundation under Grant No.QY19E050003, the Key Laboratory fund general project under Grant No.6142110190406, the key open project of 32 CETC under Grant No.22060207026, Zhejiang Education Department General Scientific Research Project under Grant No.Y201738340.(\textit{Corresponding author: Gangyong Jia.})}
\thanks{Ouhan Huang and Gangyong Jia are with The School of Computer Science and Technology, Hangzhou Dianzi University, Hangzhou, China(E-mail:huangouhan@hdu.edu.cn;gangyong@hdu.edu.cn).}
\thanks{Huanle Rao is with The School of Automation, Hangzhou Dianzi University, Hangzhou, China(E-mail:hlrao@hdu.edu.cn).}
\thanks{Xiaowen Cai is with The China Mobile Information Technology Co., Ltd., Shenzhen, China(E-mail:shiny\_c@yeah.net).}
\thanks{Ouhan Huang, Tianyun Wang, Aolong Sun , Sizhe Xing and Yifan Sun are with The Key Laboratory for Information Science of Electromagnetic Waves (MoE), Fudan University, Shanghai, China(E-mail:ohhuang23@m.fudan.edu.cn;wangtianyun23@m.fudan.edu.cn;alsun22@\\m.fudan.edu.cn;szxing21@m.fudan.edu.cn;23210720243@m.fudan.edu.cn.)}
}

\markboth{Journal of \LaTeX\ Class Files,~Vol.~14, No.~8, August~2021}%
{Shell \MakeLowercase{\textit{et al.}}: A Sample Article Using IEEEtran.cls for IEEE Journals}


\maketitle

\begin{abstract}
Forecasting vehicle behavior within complex traffic environments is pivotal within Intelligent Transportation Systems (ITS). Though this technology plays a significant role in alleviating the prevalent operational difficulties in logistics and transportation systems, the precise prediction of vehicle trajectories still poses a substantial challenge. To address this, our study introduces the Spatio Temporal Attention-based methodology for Target Vehicle Trajectory Prediction (STATVTPred). This approach integrates Global Positioning System(GPS) localization technology to track target movement and dynamically predict the vehicle's future path using comprehensive spatio-temporal trajectory data. We map the vehicle trajectory onto a directed graph, after which spatial attributes are extracted via a Graph Attention Networks(GATs). The Transformer technology is employed to yield temporal features from the sequence. These elements are then amalgamated with local road network structure maps to filter and deliver a smooth trajectory sequence, resulting in precise vehicle trajectory prediction.This study validates our proposed STATVTPred method on T-Drive and Chengdu taxi-trajectory datasets. The experimental results demonstrate that STATVTPred achieves 6.38\% and 10.55\% higher Average Match Rate (AMR) than the Transformer model on the Beijing and Chengdu datasets, respectively. Compared to the LSTM Encoder-Decoder model, STATVTPred boosts AMR by 37.45\% and 36.06\% on the same datasets. This is expected to establish STATVTPred as a new approach for handling trajectory prediction of targets in logistics and transportation scenarios, thereby enhancing prediction accuracy.
\end{abstract}

\def\abstractname{Note to Practitioners}
\begin{abstract}
This article
is motivated by the need for a high-precision trajectory prediction method for unmanned aerial vehicles (UAVs) in complex traffic scenarios. Especially in scenarios such as urban canyons or when tracking a target vehicle is lost, predicting the movement trajectory of the UAV with high precision and speed generates significant interest. In practice, as the trajectory of a target movement isn't always known, and solely relying on satellite positioning is limited by the application scenario. It presents a challenge to determine the movement trajectory of the lost UAV in a short period of time. In response to these issues, we integrated GPS positioning technology, extracted spatial attributes through a graph attention network, and used Transformer technology to extract time features from a sequence. By combining these two kinds of attributes, we obtained comprehensive spatiotemporal trajectory data, thereby dynamically predicting the future path of the UAV. The target trajectory prediction technology developed in this paper can be expanded to multiple UAV trajectory prediction scenarios to achieve coordinated path optimization, thereby improving overall performance. This paper presents an innovative approach that would benefit practitioners in the field by providing a solution to the challenging problem of high precision trajectory prediction in complex traffic scenarios.
\end{abstract}


\section{Introduction}
\IEEEPARstart{T}{he} rapid development of the global economy and population growth have increased demands for mobility, public transportation, and other civic services, exacerbating urban issues like safety, environmental pollution, and congestion\cite{kiunsi2013review}. In response, Intelligent Transportation Systems (ITS) have emerged, using advanced automation technologies such as sensors, communication networks, and AI-based decision-making modules to improve urban transport efficiency, safety, and sustainability\cite{asadi2021evolutionary,zhang2011data,lian2020review}.These technologies enable ITS to support the development of autonomous vehicles and address key urban challenges, including traffic incidents and congestion\cite{arthurs2021taxonomy,hahn2019security,ashraf2020novel}. By fostering a dynamic synergy among vehicles, infrastructure, and users, ITS pave the way for more intelligent and interconnected urban mobility, significantly alleviating traffic congestion and reducing traffic accidents.

As an essential part of the intelligent transportation system, autonomous driving technology has been widely researched and applied with the rapid development of artificial intelligence, big data, the Internet of Things(IoT) and other technologies in recent years\cite{menouar2017uav,guerrero2018sensor}. Vehicle trajectory prediction is a pivotal technology in autonomous driving, supporting critical functions in both self-driving vehicles and traffic management systems by forecasting future vehicle movements\cite{mozaffari2020deep}. This capability enables real-time decision-making and promotes safe, coordinated interactions among vehicles, pedestrians, and infrastructure. High-accuracy trajectory prediction enhances safety by allowing autonomous vehicles to anticipate and avoid potential collisions\cite{bauer2021leveraging}. Additionally, it optimizes traffic flow in congested areas through predictive re-routing and efficient fleet coordination, while also contributing to fuel savings and energy efficiency via optimized routing for delivery and commercial vehicles. Collectively, these advancements drive the operational efficacy and sustainability of modern transportation systems\cite{hegde2020velocity,kong2024exploring,kabir2023time}.

The autonomous vehicle in the Internet of Vehicles(IoV) realizes the collection of its own environment and state information using GPS, Light Detection And Ranging(LiDAR), cameras and other sensing devices. Through integrating vehicle position, speed, route and other information to build an interactive network, the use of computer technology to realize the optimal driving path selection of vehicle driving\cite{bimbraw2015autonomous,kim2019unmanned,verfuss2019review}. With the breakthroughs in critical technologies such as route navigation, obstacle avoidance and sudden decision-making, driverless technology has made breakthrough progress. It enables cars to drive autonomously without human intervention. The technology can impact traffic safety, transportation efficiency and environmental protection significantly\cite{teoh2017rage}. In addition to Waymo, other renowned companies like Uber, Tesla, Apple, Baidu and Ali have demonstrated substantial commitment by investing substantial resources in research and development efforts related to driverless technology\cite{ni2020survey}. With continuous research and development, unmanned vehicle technology is becoming commercialized and popular, making Intelligent Transportation Systems more complete and reliable.\cite{oyekanlu2020review}.However, more accurate decision-making on vehicle trajectories can better collaborate with edge-side devices and improve intelligent transportation systems.

Tracking moving targets has become one of today's research hotspots\cite{khan2016cooperative}. Target tracking, a multifaceted task comprising target detection and target tracking phases, aims to achieve real-time monitoring of a moving target's position and trajectory within video or image sequences. Unmanned vehicles rely on their camera, radar, and other sensors to gather external information, which is subsequently analyzed and identified through image recognition and various algorithms\cite{ponda2009trajectory,gao2021robust,tisdale2009autonomous}. Based on the target's motion state, corresponding decisions and planning are formulated, culminating in the effective monitoring and tracking of moving vehicles and pedestrians.In unmanned vehicle target tracking tasks, prevalent techniques include traditional visual methods and deep learning-based approaches. The conventional methods, rooted in feature-based methodologies like feature matching, correlation filtering, and particle filtering\cite{blackman1999design,bay2006surf,isard1998condensation}, coexist with deep learning-based methods that harness deep neural networks to comprehend target representations and motion patterns\cite{long2015fully,redmon2016you}. These techniques collectively contribute to the continuous evolution and improvement of target tracking capabilities in unmanned vehicle systems.

The vehicle trajectory prediction method is vital for effectively retrieving moving targets from surveillance by forecasting their future trajectories based on historical data. This process includes both traditional machine learning and deep learning techniques. Traditional methods like Support Vector Machines (SVMs), Decision Trees (DTs), and Random Forests utilize historical trajectories and environmental data to forecast future movements of target vehicles. While these methods are accurate, they require manual feature engineering and do not scale well in complex scenarios\cite{lefevre2014survey,qiao2014self}. On the other hand, advancements in deep learning have led to the adoption of methods that use Convolutional Neural Network (CNN), Long Short-Term Memory (LSTM) networks, and Self-Attention Networks (SAN) to model and predict vehicle trajectories\cite{kim2017probabilistic,deo2018convolutional,tang2018personalized,capobianco2021deep,huang2019attention}. These deep learning approaches learn features automatically, eliminating the need for manual feature creation and generally achieving higher prediction accuracy. However, they predominantly focus on short-term location forecasting and often neglect the inter-connectivity between different road sections, leading to reduced prediction reliability in those contexts\cite{9043491,9629362,kang2018self,lian2020geography,liang2021nettraj,geng2023physics}. Efforts to overcome these limitations and enhance vehicle trajectory prediction methods are ongoing, with research focusing on long-term sequence prediction and incorporating the connectivity of road sections to improve overall prediction accuracy and reliability.

To solve the above problems, this paper proposes a Spatio Temporal Attention-based methodology for Target Vehicle Trajectory Prediction (STATVTPred). First, the road segments in the road network are represented as directed graphs, and the vehicle trajectories are mapped into a sequence of road segments in the road network. The series is divided and inputted into the model. Then, the embedding layer maps the road segments into spatial vectors, and the input road segments and neighboring road segments constitute the local road network graph. The spatial features in the road network are obtained through the graph-attention network. Transformer assigns different weights to the input sequences and extracts their temporal features. Finally, combined with the local road network structure map, the output sequence is filtered to obtain a continuous trajectory sequence for vehicle trajectory prediction. The experimental results demonstrate the superior performance of STATVTPred over Transformer and LSTM Encoder-Decoder models, highlighting its effectiveness for complex trajectory prediction tasks.

Our contributions can be summarized as follows: 
\begin{itemize} 
\item We implemented a map-matching procedure to align the motion target trajectories with the road network, followed by transforming the resultant road segment sequences into feature vectors through an embedding layer.

\item The Transformer's encoding module is used to capture temporal features within the data, while the graph attention module extracts spatial features. Both modules employ the self-attention mechanism to effectively distribute dependency weights across various components.

\item The output of the Transformer's decoding module is processed through a Filter layer, resulting in a continuous sequence of predicted road segments, essential for achieving accurate target trajectory prediction.
\end{itemize}

In Section II, we review the work related to trajectory prediction for moving targets. In Section III, we outline the rating criteria for trajectory prediction and provide an introduction to the principle. In Section IV, we detail our experimental procedure and experimental analysis, including ablation experiments to demonstrate the robustness of our model. Finally, we summarize our conclusions in Section V.

\section{Related Work}
Internet of Vehicles(IoV) is a typical application of Internet of Things technology in intelligent transportation systems, which is an essential field for the deep integration of informatization and industrialization\cite{menouar2017uav}. In the context of IoV scenarios, the dynamic orchestration of vehicular movement is governed by the traffic environment\cite{ali2021machine}. The pursuit of accurate vehicle trajectory prediction has emerged as a pivotal research domain intersecting autonomous driving, dispatching systems, and vehicular safety within the IoV framework. The attainment of precise vehicle trajectory prognostication can substantially enhance the efficacy of intelligent transportation systems. In this regard, researchers have made many studies, and the existing research methods can be broadly categorized into three categories: physical model-based, statistical method-based, and machine learning-based. Some forms fuse vehicle trajectory prediction with data acquired by LiDAR and cameras to overcome the limitations of a single method. Vehicle trajectory prediction mainly includes vehicle trajectory representation and vehicle trajectory prediction.

In the modeling process, vehicle trajectory representation is the key to the mathematical abstraction of trajectory data. Vehicle trajectory representation based on spatial division is a technique to map vehicle trajectory data to discrete spatial units or regions\cite{ebel2020destination}. However, this method cannot represent the characteristics of road network topology, and significant errors will occur when dealing with precise vehicle trajectories. The road segment-based vehicle trajectory representation is a technique that matches the state of vehicle motion with small road segments\cite{tang2018personalized}. This method can obtain a more accurate form of vehicle motion, but the division of road segments needs to be reasonably designed. Intent-based vehicle trajectory representation method is a technique that correlates vehicle trajectory data with vehicle behavioural intent\cite{ding2019predicting}. This method can better understand and predict the behaviour of vehicles to achieve more innovative traffic planning, autonomous driving, and other applications. Still, the complexity and uncertainty of driver behaviour need to be considered\cite{xing2021toward}.

Upon achieving a comprehensive representation of the vehicle trajectory, the subsequent phase entails the recognition and subsequent prediction of said trajectory. Presently, prevailing vehicle trajectory prediction methodologies predominantly revolve around two foundational paradigms: probabilistic-based models and deep learning-based models. Probabilistic-based models conventionally harness statistical tools such as Markov models, Bayesian models, and Gaussian mixture models\cite{10105531,krumm2006real,xue2013destination,qiao2014self}. Through rigorous analysis of historical data, these models unveil the underlying statistical characteristics governing vehicular behavior. This, in turn, facilitates the extrapolation of future courses of action by extrapolating from historical trajectories and movement patterns. However, it's imperative to acknowledge that vehicular trajectory data within traffic scenarios tend to inhabit high-dimensional spaces, accentuating the complexity of implementing probabilistic model prediction techniques.Conversely, deep learning-based models have surfaced as potent alternatives, adept at capturing intricate temporal relationships, non-linear dynamics, and interaction patterns from expansive sequential datasets\cite{rossi2019modelling,chen2020cem}. By extracting salient features from a vehicle's historical driving state and encompassing traffic conditions, this paradigm anticipates forthcoming trajectories. Its prowess lies in its ability to enhance prediction accuracy via the assimilation of complex dependencies within trajectory data. In the pursuit of fortifying model robustness, recent research initiatives have endeavored to heighten the precision of vehicle trajectory prediction by assimilating traffic rule constraints, encompassing elements such as traffic signals and lane speed limits. Moreover, an array of studies has amalgamated deep learning-based models with their probabilistic counterparts, ushering forth more precise and reliable prediction outcomes\cite{phan2020covernet,song2020pip,sheng2022graph}. This amalgamated methodology harnesses cutting-edge models, including convolutional neural networks and recurrent neural networks. Leveraging historical vehicle trajectories, these models efficaciously prognosticate the future location and motion states of vehicles.

Although the above studies have been effective in modeling sequence data, they have yet to consider the spatial correlation of road networks fully. In urban road networks, the topology and spatial layout of the roads have an essential impact on vehicle trajectories. Therefore, if this correlation is not considered, the model's performance in predicting continuous trajectory sequences is prone to limited or even significant errors. For this reason, trajectory prediction methods that consider the spatial correlation of road networks need to be further explored to improve the accuracy and practicality of model predictions\cite{huang2022survey,leon2021review}.

\section{Principle of STATVTPred}
In this section, we first model and define the vehicle trajectory prediction problem, then introduce our modeling framework, and finally, detail the algorithmic implementation of STATVTPred.

\subsection{Problem Definition}
By representing the road sections in the road network as nodes in the directed graph and whether two road sections are connected as edges in the directed graph, the topology of the road network can be better represented. With the complete topology, the directed graph can better extract the spatial features. Finally, after we map the vehicle trajectories on the map, the prediction of the position information of the moving target is transformed into the prediction of the continuous road segment sequence.
\begin{figure*}[htb]
    \centering
    \includegraphics[width=0.9\linewidth]{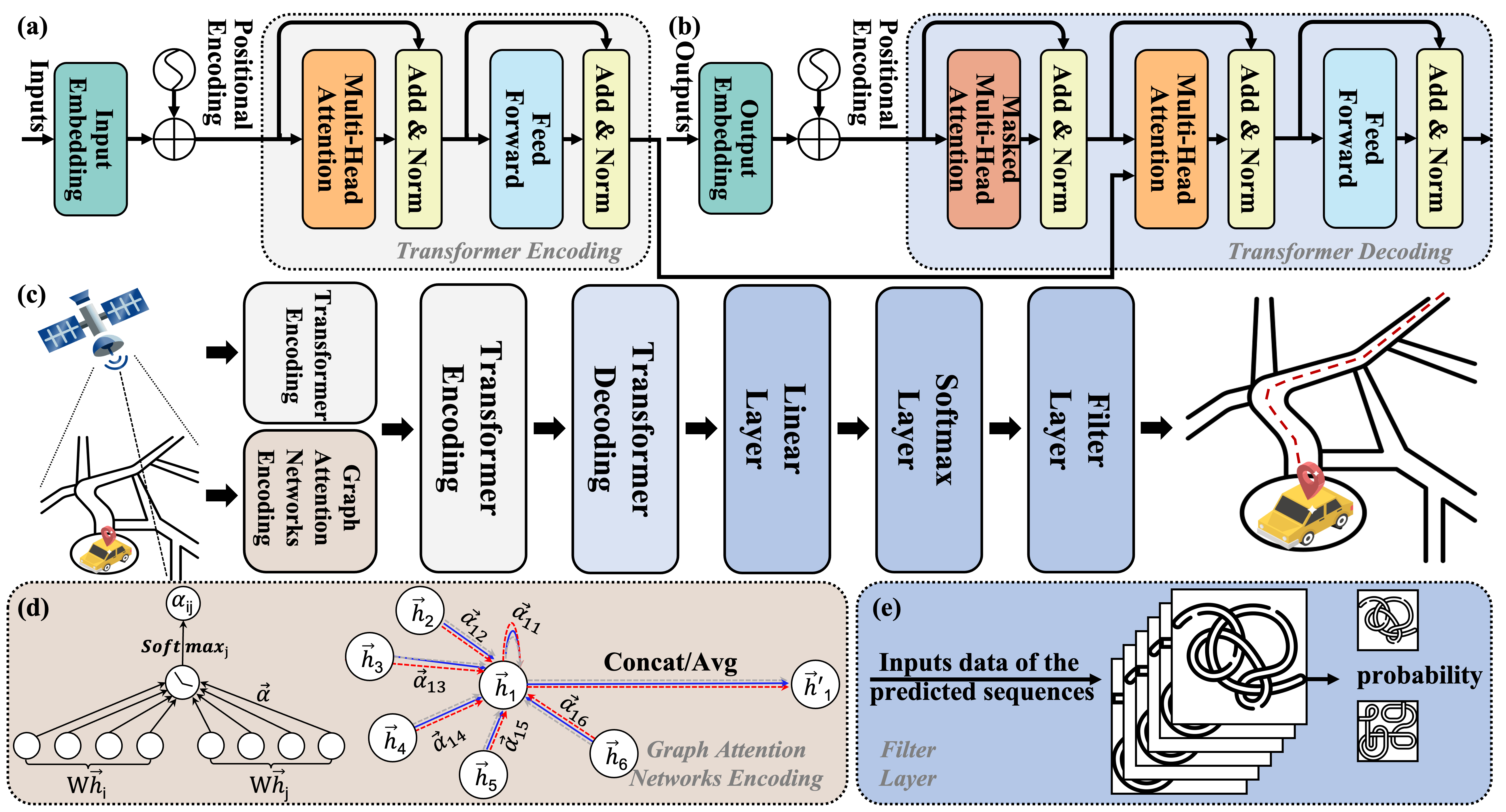}
\caption{Network structure and data handling processes. (a) is the structure of Transformer encoder, (b) is the structure of Transformer decoder, (c) is the network structure of this paper, (d) is the graph attention network structure, and (e) is the data processing process of filter layer.}
\label{fig:network}
\end{figure*}

We have implemented The following few definitions to accomplish our tasks better:

\begin{itemize}
\item Road network: the road network is represented as a directed graph $G=(V, E)$, with each road segment as a node $v\in V$ in the directed graph, $V$ denotes the set of road segments in the road network, neighboring road segments are $e\in E$, and $E$ denotes the connectivity between roads.

\item Trajectory representation: the trajectory $T$ is represented as a time-ordered sequence of consecutive sections of length $n$. $T=\left\{v_1, v_2, v_3, \ldots, v_n\right\}, \quad v_i \in V, \quad 1 \leq i \leq n, \quad<v_i, v_{i+1}>\in E$

\item Problem definition: vehicle trajectories are mapped onto a road network via a map. The ordered road sections visited by vehicles are denoted as $T=\left\{v_1, v_2, v_3, \ldots, v_{L_{i n}}\right\}$, and $L_{i n}$ is the length of the sequence of visited trajectories. The predicted ordered road section $\mathrm{P}=\left\{v_{L_{i n}}, v_{L_{i n}+2}, v_{L_{i n}+3}, \ldots, v_{L_{i n}+L_{o u t}}\right\}$ visited by the future vehicle, $L_{out}$ is the length of the predicted trajectory, and $I$ and $P$ form a continuous ordered trajectory path.
\end{itemize}

\subsection{Modeling Framework}
In the data preprocessing part of our model, we clean the GPS data to remove noise and outliers before sampling and compressing the data to retain the vital information in the data. After the data cleaning, the GPS data is mapped to the map roads by map matching algorithm and the road sequences are divided. Then, the spatial features in the trajectory data are extracted by the Graph Attention coding layer, and the Transformer coding layer is used to remove the temporal features in the trajectory data. Then we input the extracted temporal and spatial features into the Transformer coding layer after simple aggregation to further feature acquisition for such long sequences of trajectories. Finally, we use the Transformer decoder to generate the predicted sequences.

\subsubsection{Transformer module}Transformer is a neural network based on a self-attentive mechanism. Because of its powerful parallel computation capability, processing the input sequences sequentially in chronological order is unnecessary to discover the relationship between different features still. The encoder and decoder of the Transformer consist of several attentional and all-connecting layers. The attentional layer will simultaneously compute the attention weights of the input sequences to obtain a position vector representing the importance. As shown in Fig. \ref{fig:network}, the Transformer encoding layer mainly consists of a multi-head attention layer, a fully-connected feed-forward neural network layer, residual connectivity, and a normalization layer. The multi-head self-attention layer contains multiple self-attention modules that can extract the features of different attention heads in the model. The residual connection is implemented to reduce the model complexity and prevent gradient vanishing and network degradation. The normalization layer normalizes each feature's input to prevent slow model convergence due to non-uniform parameters. As shown in Fig. \ref{fig:network}, the structure of the Transformer decoder is similar to that of the encoder, with the difference that the first poly-attention of the decoding layer handles the input sequences using masks to ensure that the predicted outputs can only rely on known sequences.

The Self-Attention Mechanism (SAM) is a fundamental component of the Transformer model, enabling it to effectively capture dependencies within a sequence. SAM achieves this by computing attention weights for each position in the input sequence, thereby emphasizing the significance of different time points, which is crucial for modeling temporal data such as trajectory information. Temporal dependency capture is especially vital for trajectory prediction tasks. Leveraging the Transformer’s self-attention mechanism, our approach processes trajectory sequences in parallel, without requiring a fixed temporal order, allowing it to identify correlations across various time points and thereby enhancing prediction accuracy. Specifically, the encoder uses SAM to generate a comprehensive representation of temporal features, which the decoder then utilizes to incrementally output the predicted trajectory sequence. Moreover, the multi-head self-attention mechanism provides flexibility and effectiveness in capturing long-range dependencies by focusing on distinct temporal aspects of the sequence. In our implementation, the Multi-Head Attention mechanism is an integral part of the structure of the Transformer model. Multiple sets of Query vectors:$Q$, Key vectors:$K$, and Value vectors:$V$ are computationally generated for the input sequences and mapped to multiple subspaces through the multi-channel model. Linear changes in the multi-head attention structure are performed to input into the scaled dot product attention. The definition of scaled dot product attention is shown below:
\begin{equation}
\operatorname{Attention}(Q, K, V)=\operatorname{softmax}\left(\frac{Q K^T}{\sqrt{d_k}}\right) V,
\end{equation}
where $Q$ and $K$ are vectors of dimension $d_k$, and the dimension of $V$ is $d_v$, $Q$ and $K$ are divided by $\sqrt{d_k}$ after realizing the dot product computation, which reduces the sensitivity to the vectors' dimension and maintains the gradient's stability in training. Finally, the Softmax function is used to obtain the weights of the value vector $V$.

Multi-head Attention allows the model to learn relevant information features in different representation subspaces. The Multi-head Attention mechanism generates $d_v$ dimensional output values after h times of computation, mapping query vectors, key vectors and value vectors with different linear variations. These values are spliced and linearly transformed again to obtain the final values. This can be expressed as the Eq. \ref{eq2}

\begin{equation}
\label{eq2}
\operatorname{MultiHead}=\operatorname{Concat}\left(\text {head}_1, \text {head}_2, \ldots, \text {head}_h\right) W^O ,
\end{equation}
\begin{equation}
\text {head}_i=\operatorname{Attention}\left(Q W_i^Q, K W_i^K, V W_i^V\right),
\end{equation}
In order to understand the above equation more intuitively, we denote the set of real numbers by $R$. where $R^{a\times d}$ is denoted as a matrix of size $a\times b$ whose elements all belong to the set $R$ of real numbers. Where, $W_i^Q\in R^{d_{model}\times d_k}$,$W_i^K\in R^{d_{model}\times d_k}$,$W_i^V\in R^{d_{model}\times d_v}$ and $W^O\in R^{h d_v\times d_{model}}$, $d_{model}$ is the output dimension.In each head, the parameter matrix $W_i^Q,W_i^K,W_i^V$maps $Q,K,V$ to h subspaces under different dimensions, thus learning parameters under different features.

\subsubsection{Graph Attention Module}
Graph Attention Networks (GATs) are neural networks specifically designed to process graph-structured data by using an attention mechanism to update node feature representations and thus determine the importance of neighboring nodes to each node in the graph. Unlike traditional graph convolutional networks that treat all neighbors equally, GATs assign different attention weights to different neighbors, allowing the network to focus on the most relevant nodes when updating their representations. The Graph Attention Networks eliminates the need for complex processes such as feature decomposition or inverse of matrices. It can learn the dynamic weights between neighboring nodes without priori knowledge. The input to the graph attention layer is the set of feature representations of all nodes in the graph as follows:
\begin{equation}
\label{eq4}
\text H=\left\{h_1, h_2, h_3, \ldots, h_N\right\} \quad h_i \in R^F,
\end{equation}
Where $N$ denotes the number of nodes and $F$ denotes the number of features of a node. After the input node features are passed through the graph attention layer, the feature values of the nodes are updated and a new enhanced set of node features is obtained: 
\begin{equation}
\label{eq1}
\text H^{'}=\left\{h_1^{'}, h_2^{'}, h_3^{'}, \ldots, h_N^{'}\right\} \quad h_i^{'} \in R^{F^{'}},
\end{equation}

where $F^{'}$ denotes the number of nodes of the output node.

The graph attention layer takes the input node features and obtains new node features by a learnable linear transformation $W$, where $W \in R^{F^{'}\times F}$ is a shared linear transformation weight matrix applied to each node. The weight matrix $W$ is parameterized and applied to each node. Then the shared attention mechanism $attens:R^{F^{'}} \times R^F \rightarrow R$ is executed on these nodes and computed to obtain the attention coefficient, which is the weight coefficient between two nodes. The attention coefficient is calculated as shown in Eq. \ref{eq5}.
\begin{equation}
\label{eq5}
\text e_{ij}=attens(Wh_j,Wh_j),
\end{equation}

where node $j \in N_i$ is a neighbor node of node $i$, and $N_i$ denotes the set of neighbor nodes that have a direct relationship with node $i$ (first-order domain). The attention coefficient $e_{ij}$ reflects the importance of the features of node $j$ compared to node $i$.In the graph attention layer of the Graph Attention Network, the structural information of the graph is injected into the attention mechanism by masked attention, which fully takes into account the graph's structural information and improves the model's efficiency. Next, the attention coefficients are normalized using the Softmax function. Attention $\alpha$ is a single-layer feed-forward neural network, parameterized by the weight vector $a \in R^{2F}$. LeakyReLU is applied to introduce the nonlinearity to obtain the coefficients calculated by the attention mechanism. The calculation process is shown in Eq. \ref{eq6}.
\begin{equation}
\label{eq6}
\alpha_{i j}=\frac{\exp \left(\operatorname{LeakyReLU}\left(a^T \cdot\left[W h_i \| W h_j\right]\right)\right)}{\sum_{k \in N_i} \exp \left(\operatorname{LeakyReLU}\left(a^T \cdot\left[W h_i \| W h_k\right]\right)\right)}
\end{equation}
where $\|$ denotes the concatenation operation of vectors, and $(\cdot)^T$ represents the substitution operation of vectors.After completing the calculation of the normalized attention coefficients, the attention coefficients are weighted and summed with the corresponding node feature vectors to obtain a new feature representation for each node, as shown in Eq. \ref{eq7}:
\begin{equation}
\label{eq7}
h_i^{\prime}=\sigma\left(\sum_{j \in N_i} \alpha_{i j} \cdot W h_j\right),
\end{equation}

Repeat the operation of Eq. \ref{eq7} for M times, and perform the join operation on the features obtained through the graph attention layer to obtain the feature representation output by the multi-head attention, which is calculated as shown below:
\begin{equation}
h_i^{\prime}(M)=\prod_{m=1}^M \sigma\left(\sum_{j \in N_i} \alpha_{i j}^m \cdot W^m h_j\right),
\end{equation}
where $\alpha_{i j}^m$ denotes the attention coefficients obtained after normalization computed by the $m^{t h}$ attention mechanism, and $W^m$ represents the weight matrix undergoing linear transformation. After multi-head attention processing, the feature representation of $h_i^{'}$ contains ${MF}^{'}$ feature numbers related to the number of head in the multi-head attention mechanism.

\subsection{Implementation of The STATVTPred Algorithm}
Our algorithm implementation can be divided into a motion trajectory preprocessing phase, a spatio-temporal feature extraction phase, and a trajectory prediction phase. Then I will introduce the algorithm implementation of STATVTPred in detail.

After acquiring the motion target trajectory data via GPS, the data that do not meet the experimental requirements, such as anomalous data with too much deviation and data with too short trajectory time, are eliminated. A fast map matching algorithm combining Hidden Markov Model and pre-computation is used, which takes the GPS observation data and the road network as inputs, pre-computes all the shortest path pairs within a specific length in the road network, and stores them in the Upper Bounded Origin Destination Table(UBODT). The UBODT holds the most straightforward starting node $n_o$ to the destination node $n_d$ path, next node $next\_n$, next edge $next\_e$, previous node of the target node$pre\_n$, and shortest path distance are stored in UBODT.

\begin{figure}[htb]
    \centering
    \includegraphics[width=0.9\linewidth]{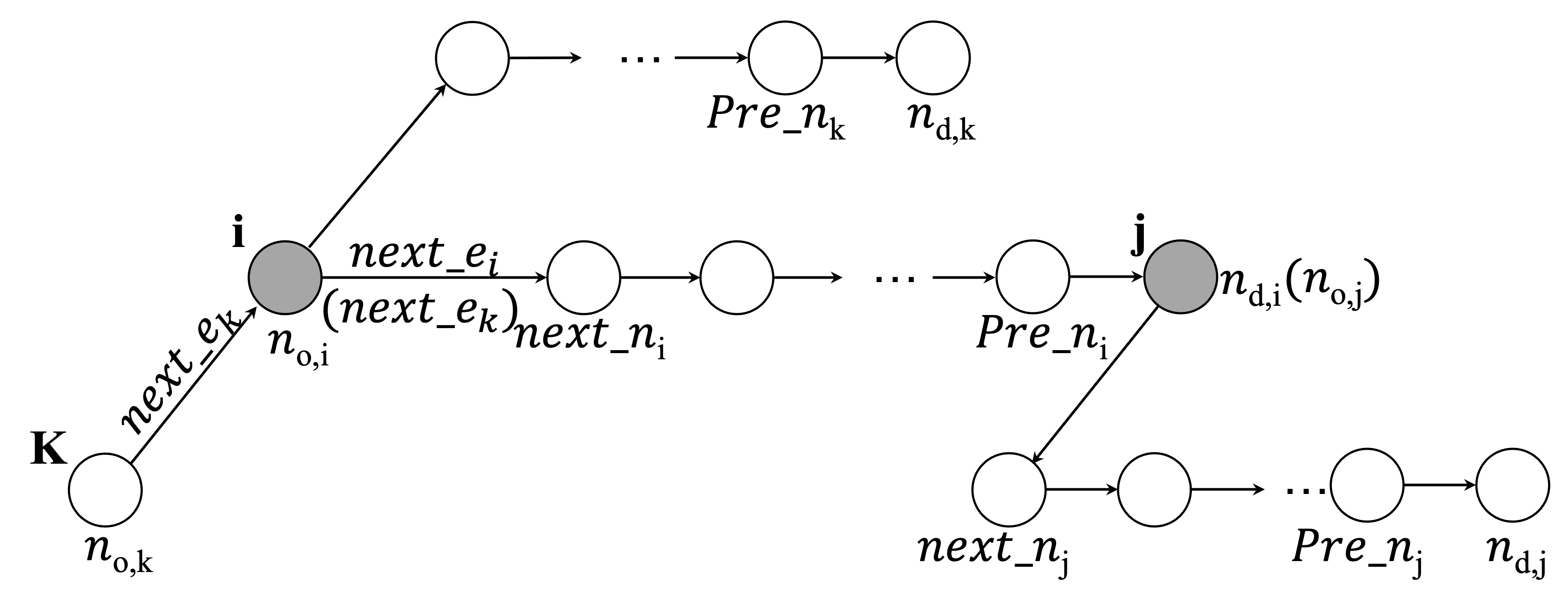}
\caption{Diagram of the process of extracting roadway neighbor information.}
\label{fig:p2}
\end{figure}

The Fig. \ref{fig:p2} shows that when the start node $n_{o,i}$ of the $i^{th}$ path is the same as the target node $n_{d,j}$ of the $j^{th}$ course, this node's front and back neighboring nodes $pre\_n_j$ and $next\_n_i$ can be obtained. When the start node $n_{o,i}$ of the $i^{th}$ path is the same as the next node $next\_n_k$ of the start node $n_{o,k}$ of the $k^{th}$ course, it can be concluded that this node's front and back neighboring edges are $next\_e_k$ and $next\_e_i$. From this, we can construct the information of every road network in the road network. Knowledge of each road segment in the road network can be built to realize the construction of the directed graph of the road network.

After preprocessing the raw trajectories, the input roadway sequence is mapped into a continuous-type feature space to facilitate the capture of feature information in the series. The road segment $v_i \in V$ in the road segment sequence is transformed into a feature vector $v_n \in R^M$, where $M$ defies the vector dimension of the road segment. The graph attention coding layer computes the weights between each node in the sequence and all its neighboring nodes to extract local spatial features, which do not depend on the complete graph structure and can handle dynamic graphs. In this paper, the road sections in the road network and their direct neighbors are used as nodes of the locally directed graph, and the graph neural network is used to obtain the spatial information of the road network. Where the input of the graph attention layer is shown in Eq. \ref{eq4}, the output is shown in Eq. \ref{eq1}, and the calculation of the attention coefficients is shown in Eq. \ref{eq5}. In the work of this paper, two layers of the attention layer are used to obtain the local road network state. The sequence nodes and their neighboring nodes are input into the graph attention coding layer. The output of the nodes and their adjacent nodes are given the weights to get the spatial relationship between the sequence nodes and their direct neighbors in this section of the road.

The Transformer encoding layer consists of two sublayers, the multi-head attention mechanism, and the feed-forward neural network. The input sequence is transformed independently by multiple attention heads. Then the outputs of all the attention heads are spliced together to generate a contextual representation of the location by a linear transformation. The self-attention mechanism implements parallel computation to encode each location in the input sequence and captures sequence dependencies to extract temporal features.

The output sequence of the final Transformer encoding layer is passed to the decoding layer for decoding. The decoding layer outputs each target road segment step by step through an internal attention mechanism, and the location information of each road segment can be determined by adding the location code in the decoding layer. After selecting the location codes, the decoding layer outputs the decoded results individually. The results of the decoding layer are processed by the Linear layer, Softmax layer, and Filter layer to output the prediction sequence. After the trajectory preprocessing stage to obtain the road network directed graph $G_{vec}$, the Filter layer filters the output data according to $G_{vec}$. It generates the neighbor vector $W_c$ after receiving the connectivity relationship, as shown below:
\begin{equation}
W_c=\left[\begin{array}{cccc}
w_{1,1} & w_{1,2} & \cdots & w_{1, n_c} \\
w_{2,1} & w_{2,2} & \cdots & w_{2, n_c} \\
\vdots & \vdots & \ddots & \vdots \\
w_{n_c, 1} & w_{n_c, 2} & \cdots & w_{n_c, n_c}
\end{array}\right], 
\end{equation}

where ${W_{i,j}=1}, {i,j} \in G_{vec}$ then indicates that the segments are connected and ${W_{i,j}=0}, {i,j} \notin G_{vec}$ indicates that the segments are disconnected. the Softmax layer outputs the probability that each segment may be the next.

\section{Experiments Evaluation}
In this section, we describe the experimental setup. We present the experimental data and the evaluation system, then analyze the experimental results and compare the baseline methods. Finally, we perform ablation experiments to demonstrate our model's robustness and generalization ability.

\subsection{Environment Configuration}

The configuration of the environment used for the experiments in this paper is as follows: the CPU is Intel(R) Core(TM) i7-7800X CPU @3.50GHz, the memory capacity is 64GB, the GPU is NVIDIA GeForce RTX 2080Ti, the operating system is Ubuntu 18.04, the deep learning framework is Pytorch and the Python modules include Numpy, Pandas and so on. Among them, the parameters of Pytorch are selected as shown in Table  \ref{tab.table1}.
\begin{table}[ht]
    \caption{Deep learning framework parameter selection in this experiment}
    \label{tab.table1}
    \centering
    \footnotesize
    \setlength{\tabcolsep}{3.8mm}{
    \begin{tabular}{lc}
        \toprule
        \specialrule{0em}{-1pt}{1pt}
        \rule[-1ex]{0pt}{3.5ex}Deep learning framework parameters & Parameter selection \\
        \hline
        \rule[-1ex]{0pt}{3.5ex}Length of the input data sequence & 8 \\
        \hline
        \rule[-1ex]{0pt}{3.5ex}Length of the output data sequence & 4 \\
        \hline
        \rule[-1ex]{0pt}{3.5ex}Optimizer & Adaptive momentum \\
        \hline
        \rule[-1ex]{0pt}{3.5ex}Learning rate & 0.5 \\
        \hline
        \rule[-1ex]{0pt}{3.5ex}Activation Function & Rectified Linear Unit \\
        \hline
        \rule[-1ex]{0pt}{3.5ex}Weight decay & 0.01 \\
        \hline
        \rule[-1ex]{0pt}{3.5ex}Batch size & 100 \\
        \hline
        \rule[-1ex]{0pt}{3.5ex}Dropout & 0.1 \\
        \hline
        \rule[-1ex]{0pt}{3.5ex}epoch & 40 \\
        \specialrule{0em}{-1pt}{1pt}
        \bottomrule
    \end{tabular}}
\end{table}

\subsection{Dataset}

\begin{figure*}[h]
    \centering
    \includegraphics[width=0.9\linewidth]{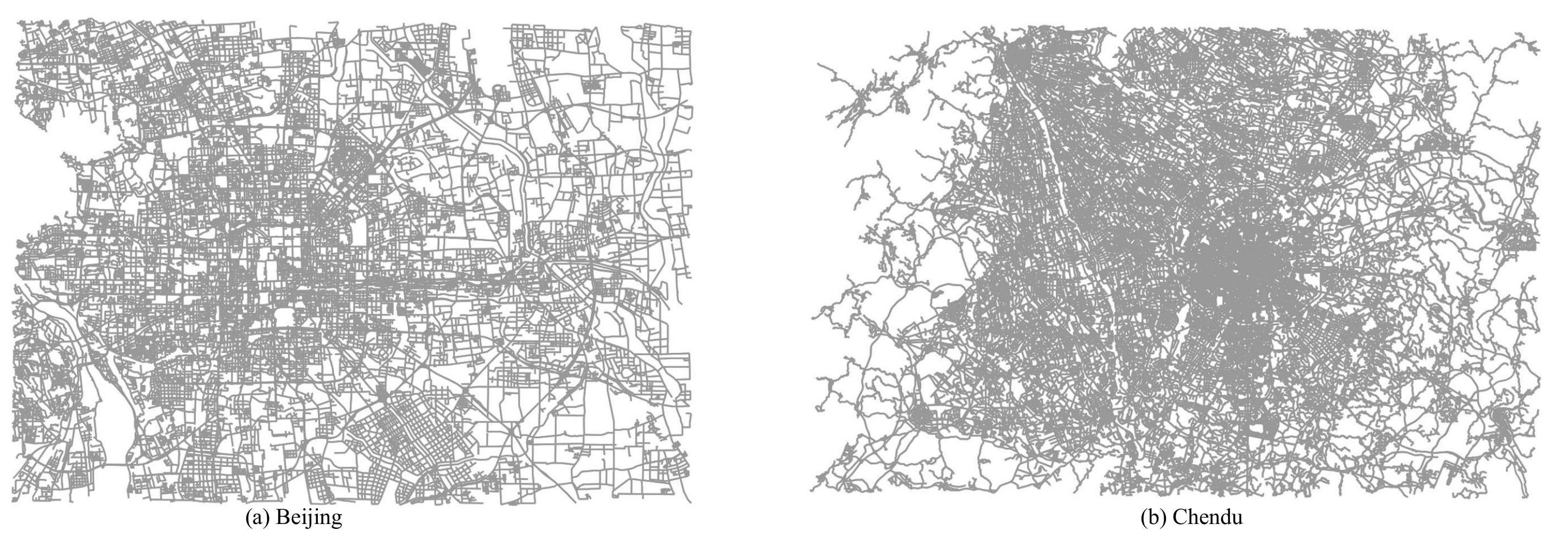}
\caption{Map of the road network used in the experiment. (a) Beijing City Road Network and (b) Chendu City Road Network.}
\label{fig:road}
\end{figure*}

To verify the performance of the model proposed in this paper, many experiments are conducted on two accurate vehicle trajectory datasets, Beijing Taxi Trajectory and Chengdu Taxi Trajectory, and the structure of the datasets is shown in Table \ref{tab.table2} and Table \ref{tab.table3}. 
\begin{table}[ht]
    \caption{Sample Beijing dataset}
    \label{tab.table2}
    \centering
    \footnotesize
    \setlength{\tabcolsep}{4mm}{
    \begin{tabular}{cccc}
        \toprule
        \specialrule{0em}{-1pt}{1pt}
        \rule[-1ex]{0pt}{3.5ex}ID & Time & Longitudes & Longitude\\
        \hline
        \rule[-1ex]{0pt}{3.5ex}1 & 2008-02-02 15:36:08 & 116.51172 & 39.92123\\
        \hline
        \rule[-1ex]{0pt}{3.5ex}1 & 2008-02-02 15:46:08 & 116.51135 & 39.93883\\
        \hline
        \rule[-1ex]{0pt}{3.5ex}2 & 2008-02-02 13:33:52 & 116.36422 & 39.88781\\
        \hline
        \rule[-1ex]{0pt}{3.5ex}2 & 2008-02-02 13:37:16 & 116.37481 & 39.88782\\
        \hline
        \rule[-1ex]{0pt}{3.5ex}3 & 2008-02-02 13:39:08 & 116.35743 & 39.88957\\
        \hline
        \rule[-1ex]{0pt}{3.5ex}3 & 2008-02-02 13:44:08 & 116.35732 & 39.89726\\
        \specialrule{0em}{-1pt}{1pt}
        \bottomrule
    \end{tabular}}
\end{table}
\begin{table}[ht]
    \caption{Sample Chendu dataset}
    \label{tab.table3}
    \centering
    \footnotesize
    \setlength{\tabcolsep}{2mm}{
    \begin{tabular}{ccccc}
    \toprule
    \specialrule{0em}{-1pt}{1pt}
        \rule[-1ex]{0pt}{3.5ex}ID & Time & Longitudes & Longitude & Situation\\
        \hline
        \rule[-1ex]{0pt}{3.5ex}1 & 2014/08/04 15:24:08 & 104.002727 & 30.575682 &0\\
        \hline
        \rule[-1ex]{0pt}{3.5ex}1 & 2014/08/04 15:24:13 & 104.002931 & 30.575637&1\\
        \hline
        \rule[-1ex]{0pt}{3.5ex}2 & 2014/08/04 12:10:18 & 104.128109 & 30.636512&1\\
        \hline
        \rule[-1ex]{0pt}{3.5ex}2 & 2014/08/04 12:10:29 & 104.128105 & 30.636506&1\\
        \hline
        \rule[-1ex]{0pt}{3.5ex}3 & 2014/08/04 14:28:35 & 103.981980 & 30.636185&0\\
        \hline
        \rule[-1ex]{0pt}{3.5ex}3 & 2014/08/04 14:29:36 & 103.981960 & 30.636197&0\\
        \specialrule{0em}{-1pt}{1pt}
        \bottomrule
    \end{tabular}}
\end{table}
The Beijing cab trajectory dataset comes from T-Drive, which contains one-week trajectories of 10,357 cabs from February 2, 2008, to February 8, 2008, in Beijing. Each data record includes the information of vehicle ID, time, longitude and latitude. The road network structure at the acquisition boundary [39.74,40.10,116.14,116.80] consists of 46859 intersection nodes and 109,314 road segments, as shown in Fig. \ref{fig:road}(a). The Chengdu cab trajectory dataset contains 11 days of data collected from 13,605 cab drivers in Chengdu from August 3, 2014, to August 13, 2014, via smartphones. Each GPS data record describes the cab carrying situation at a specific time and location, including time, vehicle ID, occupancy status, longitude and latitude. The road network structure at the acquisition boundary [30.29,31.04,103.26,104.61] consists of 83,483 intersection nodes and 203,578 road segments, as shown in Fig. \ref{fig:road}(b).

After data preprocessing, the trajectory records of the dataset were mapped onto the road network structure, and invalid data were removed. The data sequences were processed into input sequences of length eight and output sequences of length four. Among them, the processed Beijing dataset contains 124845 sets of trajectory sequences, and the Chengdu dataset contains 93482 sets of trajectory sequences. The training, validation, and test sets were randomly divided into 80\%, 10\% and 10\%, respectively.

\subsection{Experimental Evaluation Indicators }
In our experiments, we use the Distance Error(DE) and Average Match Rate(AMR) as the performance evaluation metrics of the vehicle trajectory prediction model. The DE is quantified by calculating the average edit distance between the predicted and actual trajectories. The edit distance quantifies the similarity between two sequences by calculating the minimum number of operations required to convert one sequence to another. The DE is defined as shown in Eq. \ref{eq11}:

\begin{equation}
\label{eq11}
D E=\frac{1}{N_t L_{\text {out }}} \sum_{i=1}^{N_t} \operatorname{Edit}\left(P_i^{\prime}, P_i\right),
\end{equation}
where $N_t$ denotes the number of trajectory records in the test set, $L_{out}$ denotes the length of the output trajectory, $P_i^{'}$ denotes the trajectory sequence obtained by prediction, and $P_i$ denotes the actual trajectory sequence. $Edit(\cdot,\cdot)$ denotes the edit distance between the two, i.e., the degree of difference between the two sequences is calculated. 

Average Match Rate (AMR): the average ratio of the number of correctly predicted road segments to the length of the predicted trajectory. The AMR formula is defined as: 

\begin{equation}
A M R=\frac{1}{N_t L_{o u t}} \sum_{i=1}^{N_t} \sum_{j=1}^{L_{o u t}} \operatorname{Match}\left(v_{i, j}^{\prime}, v_{i, j}\right),
\end{equation}

Where $v_{i,j}^{'}$ and $v_{i,j}$ denote the road segment predicted from the $j^{th}$ position of the $i^{th}$ trajectory sequence and the actual road segment, respectively, and $Match(\cdot,\cdot)$ denotes the degree of matching between the two. When the two are the same, ma=1, otherwise 0.

\subsection{Results and Discussion}

In this subsection, we will detail our experimental results and justify our constructed model through ablation experiments, and finally, we will give the convergence procedure of our experiments.

\subsubsection{Comparative experiments with conventional}

\begin{table}[!t]
\begin{center}
\caption{Comparison with conventional modeling experiments.}
\label{tab:ana}
\begin{tabular}{ccccc}\toprule
\multirow{2.5}{*}{Method} & \multicolumn{2}{c}{Beijing}  & \multicolumn{2}{c}{Chendu}     \\ \cmidrule(l{2pt}r{2pt}){2-3} \cmidrule(l{2pt}r{2pt}){4-5}
                          & DE(\%) & AMR(\%) & DE(\%) & AMR(\%) \\ \midrule
LSTM Encoder-Decoder& 63.81 & 35.62 & 56.32 & 42.87 \\
\midrule
Transformer& 33.27 & 66.69 & 31.52 & 68.38 \\
\midrule
STATVTPred& 26.93 & 73.07 & 20.95 & 78.93 \\
\bottomrule
\end{tabular}
\end{center}
\end{table}
In this paper, LSTM Encoder-Decoder and Transformer are used as the baseline models, and the performance of the proposed method is evaluated by comparing them with the baseline models. On the Beijing dataset, the AMR value of STATVTPred reaches 73.07\%, which is 37.45\% and 6.38\% higher than that of LSTM Encoder-Decoder and Transformer, respectively. On the Chengdu dataset, the AMR value of STATVTPred reaches 78.93\%, which is 36.06\% and 10.55\% higher than that of LSTM Encoder-Decoder and Transformer, respectively. It can be observed that for both datasets, the proposed method STATVTPred has better performance than the baseline model. This is because existing deep learning methods cannot capture spatial correlation. At the same time, the model proposed in this paper uses graph structure to extract spatial features of road networks, which can capture spatial correlation more efficiently. The attention mechanism can better capture temporal correlation on time series.

\begin{table}[!t]
\begin{center}
\caption{Comparison with state-of-the-art methods on the Beijing dataset.}
\label{tab:com}
\begin{tabular}{ccc}\toprule
\multirow{2.5}{*}{Method} & \multicolumn{2}{c}{Beijing}      \\ \cmidrule(l{2pt}r{2pt}){2-3} 
                          & DE(\%) & AMR(\%) \\   \midrule
Markov Chain\cite{qiao2014self}& 34.1 & 65.8  \\
\midrule
Convolutional Sequence Embedding\cite{tang2018personalized} & 31.4 & 68.5  \\
\midrule
Attentional LSTM Encoder-Decoder\cite{capobianco2021deep} & 31.7 & 68.1  \\
\midrule
Attentional Spatiotemporal LSTM\cite{huang2019attention}& 32.8 & 67.0  \\
\midrule
Self-Attention Based Sequential Model\cite{kang2018self}& 31.5 & 68.4  \\
\midrule
Geography-Aware Self-Attention Network\cite{lian2020geography}& 33.0 & 66.9  \\
\midrule
Transformer\cite{geng2023physics}& 33.3 & 66.7  \\
\midrule
NetTraj\cite{liang2021nettraj}& 30.1 & 69.7  \\
\midrule
\textbf{STATVTPred(This Work)}& \textbf{26.9} & \textbf{73.1}  \\
\specialrule{0em}{-1pt}{1pt}
\bottomrule
\end{tabular}
\end{center}
\end{table}

\begin{figure}[htb]
    \centering
    \includegraphics[width=0.9\linewidth]{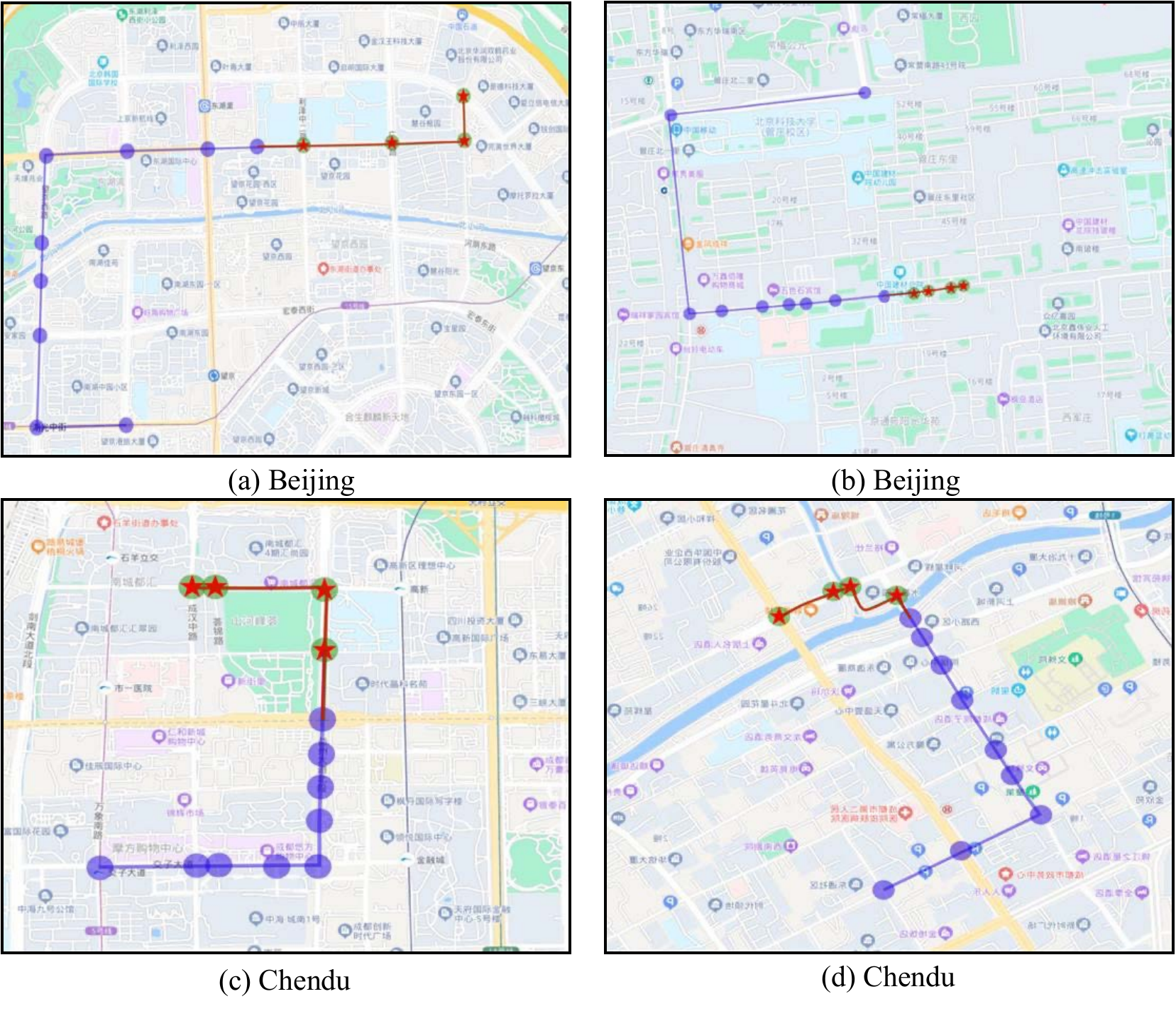}
\caption{The trajectory prediction results of STATVTPred.  (a),(b) Predicted results for the Beijing dataset, (c),(d) Predicted results for the Chendu dataset.}
\label{fig:roadPred}
\end{figure}
The STATVTPred prediction results are shown in Fig. \ref{fig:roadPred}, where the purple part indicates the historical trajectory, the green part indicates the target trajectory and the red part indicates the trajectory after model prediction. From the figure, it can be seen that the trajectory prediction results are as expected. The results of comparing our model with the baseline model are shown in Table \ref{tab:ana}.

In order to further validate the effectiveness of our proposed method, we conducted a comparative analysis with several state-of-the-art methods developed by other researchers on the Beijing dataset as a benchmark. The results show that our method achieves significantly lower error rates, highlighting its superiority over alternative methods. The detailed experimental results are listed in Table \ref{tab:com}.

The performance of the model STATVTPred proposed in this paper is compared to the Beijing dataset with input sequences of different lengths.Compare the performance of STATVTPred with different lengths of input sequences on the Beijing dataset. In this set of experiments, the input sequence length and output sequence length are gradually increased, and the input sequence length is 2, 4, 6, 8, and the output sequence length is 1, 2, 3, 4. According to the data in Table \ref{tab:tab22}, we can find that the model's performance gradually improves with the increase in the length of the input sequence. When the input length is 8, and the output length is 4, the AMR is 73.07\%. Although the increase in input sequence length can improve the performance, we can see from the data that the long-term dependent features do not contribute much to the performance improvement, and most of the informative features can be obtained from the two neighboring sections. In addition, from the data in Table \ref{tab:tab22}, it can also be found that the accuracy of the prediction gradually decreases as the length of the output sequence increases. 
\begin{table*}[!htp]
\begin{center}
\caption{Performance of different input sequence lengths on the Beijing dataset(T-Drive) }
\label{tab:tab22}
\begin{tabular}{ccccccccc}\toprule
\specialrule{0em}{-1pt}{1pt}
\multirowcell{4}{Input \\Sequence\\ Length} & \multicolumn{8}{c}{Output Sequence Length} 
\\& \multicolumn{2}{c}{1} & \multicolumn{2}{c}{2}& \multicolumn{2}{c}{3} & \multicolumn{2}{c}{4}   
\\ \cmidrule(l{2pt}r{2pt}){2-3} \cmidrule(l{2pt}r{2pt}){4-5}
\cmidrule(l{2pt}r{2pt}){6-7} \cmidrule(l{2pt}r{2pt}){8-9}
                          & DE(\%) & AMR(\%) & DE(\%) & AMR(\%) & DE(\%) & AMR(\%)& DE(\%) & AMR(\%)\\ \midrule
                          2& 14.64 & 85.36 & 20.56 & 79.44 & 25.86 & 74.14& 30.85 & 69.13\\
                          \midrule
                          4& 13.41 & 86.59 & 19.05 & 80.95 & 24.10 & 75.90& 28.61 & 71.37\\
\midrule
                          6& 13.06 & 86.94 & 18.47 & 81.33 & 23.35 & 76.34& 27.75 & 72.24\\
                          \midrule
                          8& 12.56 & 87.44 & 17.89 & 82.11 & 22.67 & 77.33& 26.93 & 73.07\\
\specialrule{0em}{-1pt}{1pt}
\bottomrule
\end{tabular}
\end{center}
\end{table*}

\begin{figure*}[htb]
    \centering
    \includegraphics[width=0.9\linewidth]{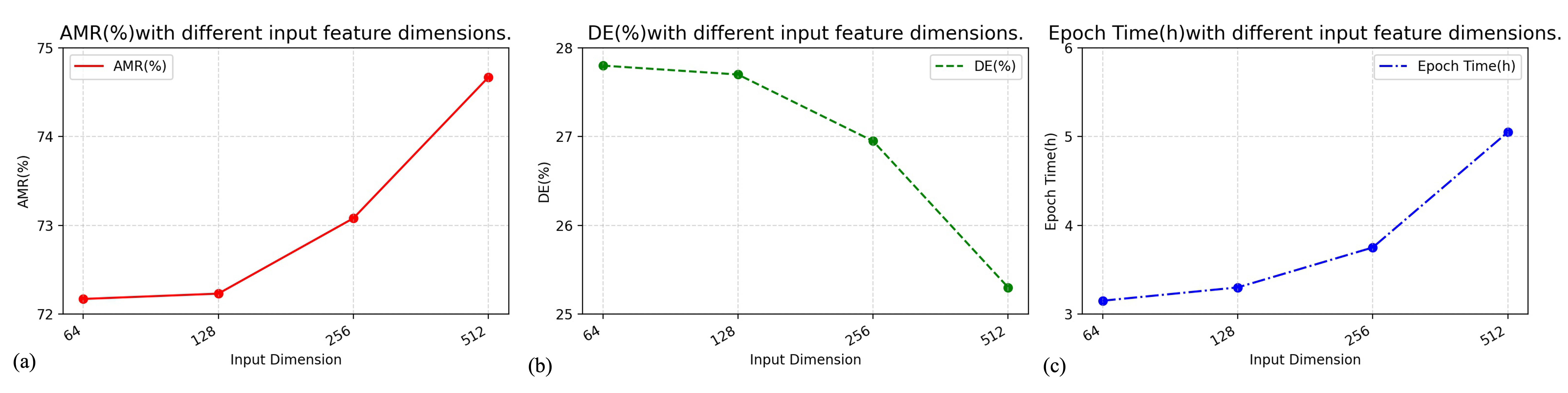}
\caption{Comparison plot of the effect of different input feature dimensions on the experiment. }
\label{fig:com}
\end{figure*}

Since different input feature dimensions will impact the model performance and training effect, this paper conducts experiments based on the Beijing dataset to determine reasonable input feature dimensions. Fig. \ref{fig:com}(a) and Fig. \ref{fig:com}(b) show the effects of different input feature dimensions on AMR and DE, respectively. Fig. \ref{fig:com}(c) represents the training time of each epoch under different input feature dimensions, i.e., the effect of varying input feature dimensions on the training time.

The overall performance of STATVTPred keeps improving as the input feature dimension increases. Higher input feature dimensions increase the capacity of the model and therefore enhance the performance of the model. However, higher feature dimensions result in the need for more parameters and calculations, and more time is needed to process and update these parameters during training Therefore, based on a comprehensive analysis of the model's performance and training time, the default value of the input feature dimension is set to 256 in this paper.

\subsubsection{Comparative convergence experiments}
\begin{figure}[htb]
    \centering
    \includegraphics[width=0.9\linewidth]{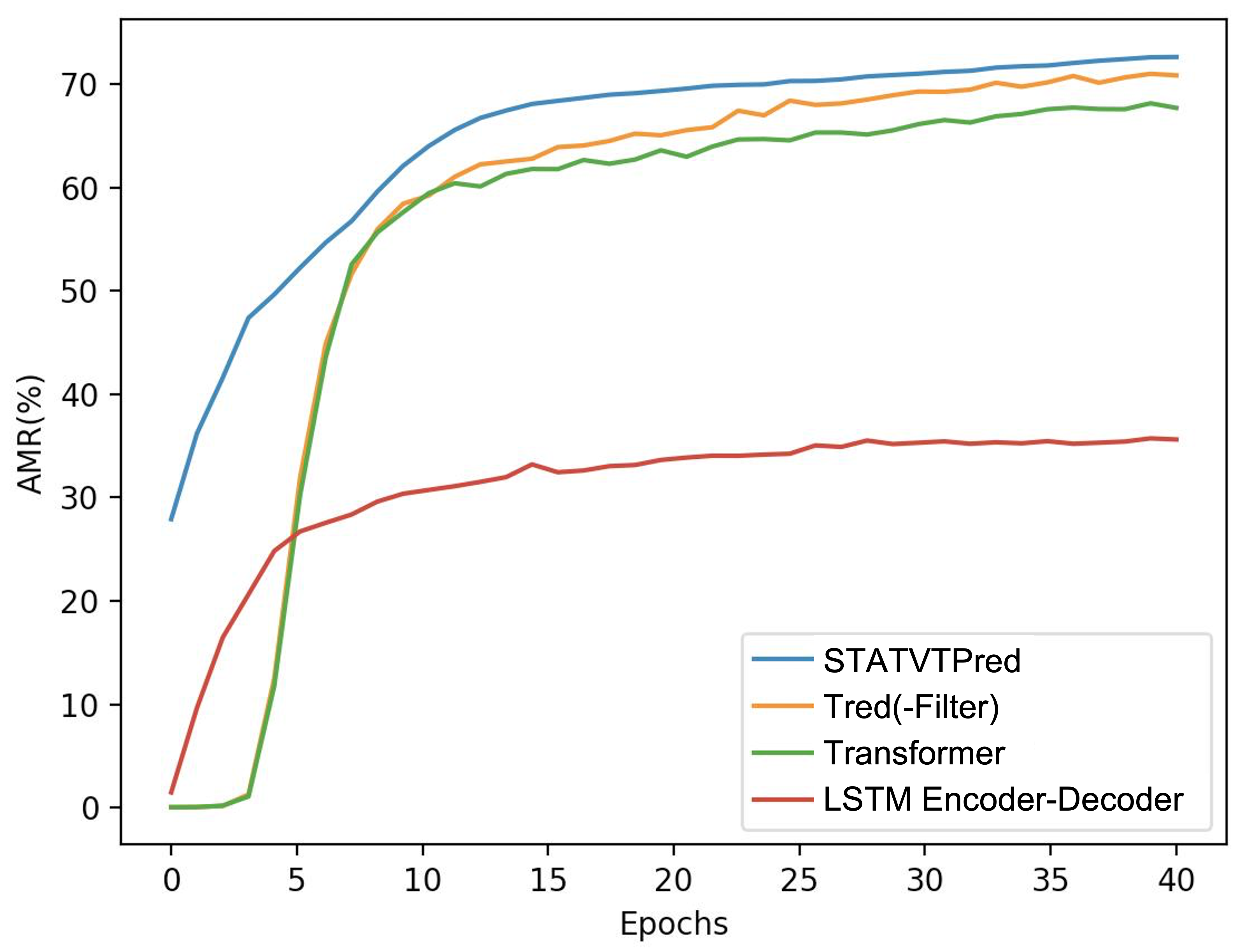}
\caption{Convergence curves on different models on the Beijing dataset(T-drive).}
\label{fig:7}
\end{figure}

Fig. \ref{fig:7} shows the convergence process of the Beijing dataset on different models. As can be seen from the figure, these models stabilize after 40 epochs, and the LSTM Encoder-Decoder model is the fastest model to reach a smooth state because of its rapid convergence speed, although its accuracy is lower. Although the accuracy of the LSTM Encoder-Decoder is scarce, its convergence speed is fast, and it is the quickest model to reach a smooth state. The Transformer model has a slower convergence speed, but its performance is significantly better than the LSTM Encoder-Decoder as time passes. After adding the Filter module, TPred outperforms the other models significantly. This is because the Filter module utilizes the connectivity between road segments to filter out disconnected road segments, effectively improving the prediction's accuracy.

\subsection{Ablation Experiments}
In this segment, To explore in depth the differences in the performance of the models. We conduct an ablation study to validate the efficacy of individual components within our model across varying conditions.

Since there are more data in Beijing than in Chengdu, this paper will focus on the trajectory data in Beijing to verify the model effect. In the model proposed in this paper, the Graph Attention Module and Filter Module are important influencing factors. We represent the model presented in this paper as STATVTPred and place different components to build variants, which are tested and compared with the full model. To validate the effectiveness of these two modules on the network model, several sets of ablation experiments are done separately to compare with the network model STATVTPred designed in this paper.
\begin{table}[!t]
\begin{center}
\caption{Ablation experiments on the Beijing dataset.}
\label{tab:duibi}
\begin{tabular}{ccc}\toprule
\multirow{2.5}{*}{Method} & \multicolumn{2}{c}{Beijing}      \\ \cmidrule(l{2pt}r{2pt}){2-3} 
                          & DE(\%) & AMR(\%) \\   \midrule
LSTM Encoder-Decoder& 63.81 & 35.62  \\
\midrule
LSTM Encoder-Decoder(+Filter)& 35.84 & 64.16  \\
\midrule
Transformer& 33.27 & 66.69  \\
\midrule
STATVTPred(-Filter)& 29.08 & 70.89  \\
\midrule
\textbf{STATVTPred}& \textbf{26.93} & \textbf{73.07}  \\
\specialrule{0em}{-1pt}{1pt}
\bottomrule
\end{tabular}
\end{center}
\end{table}

As shown in the Table \ref{tab:duibi}. The ablation experiment contains the following cases, LSTM Encoder-Decoder(+Filter) denotes that the Filter module filters the LSTM Encoder-Decoder model after the output result. STATVTPred(-Filter) means that the model in this paper removes the Filter module and does not filter the output result. STATVTPred(-Filter) compares with STATVTPred to determine the Filter module's effectiveness. Comparison of STATVTPred(-Filter) with Transformer can reflect the positive effect of the graph attention module. By these comparisons, the network effectiveness of the TPred model can be effectively demonstrated.

The ablation experiments in this paper also incorporate a comparison experiment between LSTM Encoder-Decoder and LSTM Encoder-Decoder(+Filter). The comparison results show that the AMR of the former is 35.62\% while the AMR of the latter is 64.16\%, a difference of 28.54\%. The AMR of STATVTPred and STATVTPred(-Filter) is only improved by 2.18\% compared to STATVTPred(-Filter). From the experimental results in Table \ref{tab:duibi} , it can be seen that adding the Filter module can significantly increase the accuracy of model prediction when the accuracy of model prediction is low compared to the comparison experiments of STATVTPred and STATVTPred(-Filter)—accuracy of model prediction. With the ablation experiments and the gradual recovery of the network modules, we can find that the closer to the complete network model structure, the higher the model prediction accuracy.

\section{Conclusion}
In this paper, we proposed STATVTPred, a spatio-temporal attention-based method for predicting target vehicle trajectories. STATVTPred employs a Transformer architecture to encode and decode vehicle trajectory sequences using the road network’s graph structure and an attention mechanism. This process includes filtering with a directed graph representation of the road network, allowing for accurate future trajectory predictions. During data preprocessing, GPS records are mapped to the road network, generating ordered road sequences; the graph attention module captures spatial features, while the Transformer module leverages self-attention to model temporal features.By effectively capturing complex spatial and temporal dependencies in dynamic traffic environments, STATVTPred demonstrates substantial improvements in prediction accuracy and computational efficiency across multiple datasets, particularly in long-distance and complex trajectory prediction scenarios. Extensive experiments on two real-world datasets, Beijing and Chengdu, show that STATVTPred achieves an AMR of 73.07\% on the Beijing dataset, surpassing the LSTM Encoder-Decoder and Transformer baselines by 37.45\% and 6.38\%, respectively. Similarly, on the Chengdu dataset, STATVTPred achieves an AMR of 78.93\%, outperforming LSTM Encoder-Decoder by 36.06\% and Transformer by 10.55\%. These results validate the robustness and accuracy of STATVTPred in vehicle trajectory prediction tasks.
Furthermore, STATVTPred’s integration of road network-based filtering enhances trajectory smoothness and adaptability, supporting applications such as self-driving car navigation, predictive routing, and traffic flow management, making it highly valuable for Intelligent Transportation Systems (ITS). This model contributes to safer, more efficient urban transportation. Additionally, STATVTPred provides a flexible framework adaptable to various trajectory prediction scenarios, from real-time fleet management in logistics to proactive traffic control systems. 
This work lays a foundation for the further development of intelligent traffic prediction models and supports advancements in the field of intelligent, interconnected vehicle systems.

\bibliographystyle{IEEEtran}
\bibliography{trans}

\begin{IEEEbiography}[{\includegraphics[width=1in,height=1.25in,clip,keepaspectratio]{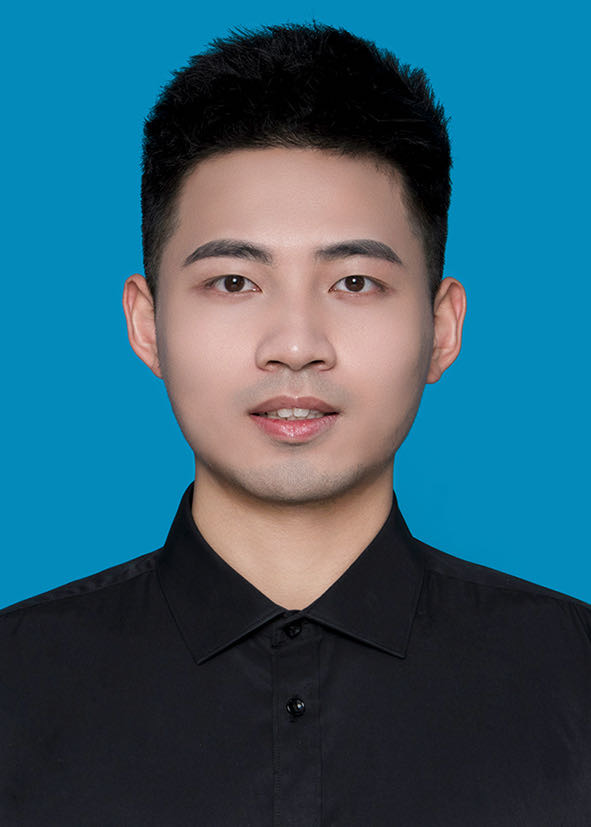}}]{Ouhan Huang}
received his B.E. degree in Internet of Things Engineering from from Hangzhou Dianzi University, Hangzhou, China in 2020, and his M.S. degree in Computer Science and Technology from ITMO University, St. Petersburg, Russia in 2022. He is currently pursuing a Ph.D. degree at the School of Information Science and Engineering, Fudan University. His research interests include Machine Learning Algorithms, Visible Light Communication and Edge Computing.
\end{IEEEbiography}
\vspace{-10mm}

\begin{IEEEbiography}[{\includegraphics[width=1in,height=1.25in,clip,keepaspectratio]{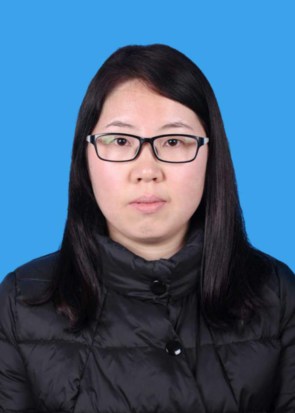}}]{Huanle Rao}
received her B.S. degree in Optical Information Science and Technology from Hefei University of Technology, Hefei, in 2008, and her Ph.D. degree in Engineering Nuclear Science and Technology from University of Science and Technology of China, Hefei, in 2014. Her research interests include edge computing and machine learning. She is a member of IEEE.
\end{IEEEbiography}
\vspace{-10mm}

\begin{IEEEbiography}[{\includegraphics[width=1in,height=1.25in,clip,keepaspectratio]{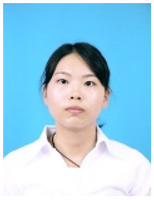}}]{Xiaowen Cai} 
received her B.S. degree from North China Electric Power University, Baoding, China in 2018, and her M.S. degree with the Department of Computer, Hangzhou Dianzi University, Hangzhou, China in 2023. Her research interests include unmanned systems and intelligent transportation systems.
\end{IEEEbiography}
\vspace{-10mm}

\begin{IEEEbiography}[{\includegraphics[width=1in,height=1.25in,clip,keepaspectratio]{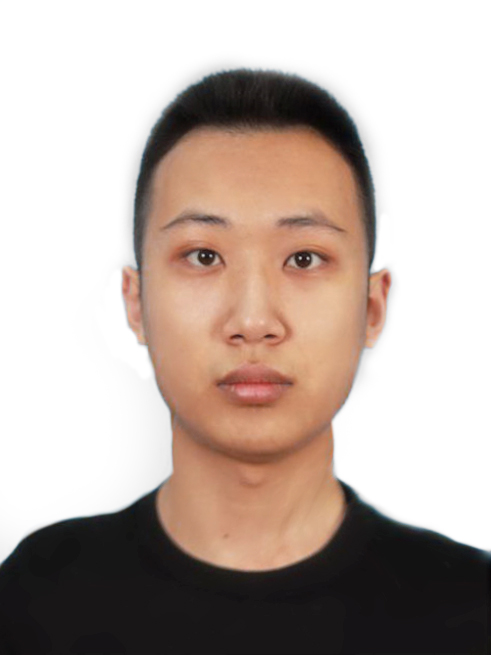}}]
{Tianyun Wang}
received his B.E. degree in school of mechatronics engineering from Harbin Institute of Technology, Harbin, China in 2023. He is currently pursuing a Ph.D. degree at the School of Information Science and Engineering, Fudan University. His research interests include Computational Imaging, Generative Models and Deep Learning Algorithms.
\end{IEEEbiography}
\vspace{-10mm}

\begin{IEEEbiography}[{\includegraphics[width=1in,height=1.25in,clip,keepaspectratio]{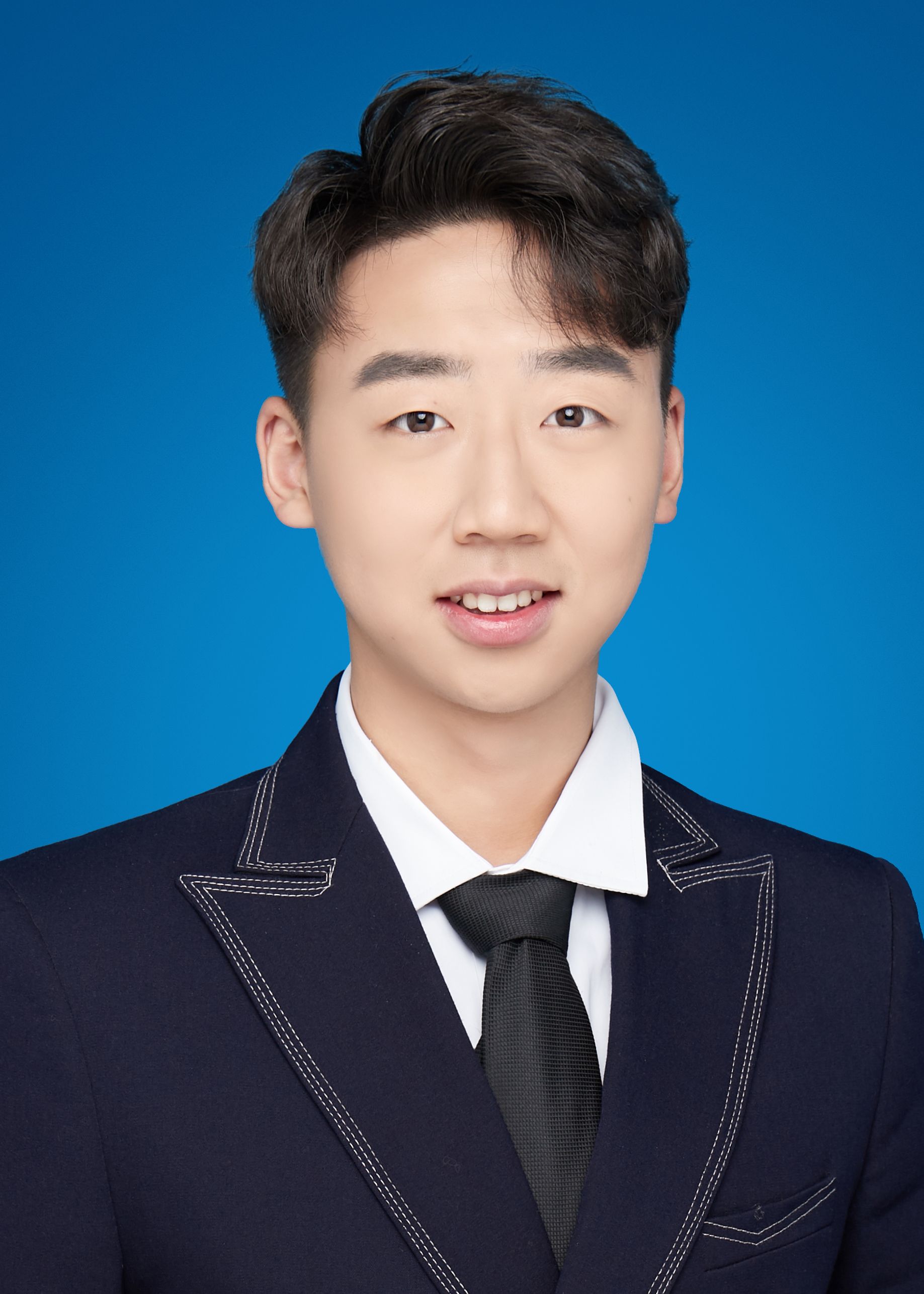}}]
{Aolong Sun}
received his B.S. degree in Optical Information Science and Technology from Huazhong University of Science and Technology, Wuhan, China in 2022. He is currently pursuing a Ph.D. degree at the School of Information Science and Engineering, Fudan University. His research interests include Silicon Photonics, Optical Computing and On-chip Optical Communication Systems.
\end{IEEEbiography}
\vspace{-10mm}

\begin{IEEEbiography}[{\includegraphics[width=1in,height=1.25in,clip,keepaspectratio]{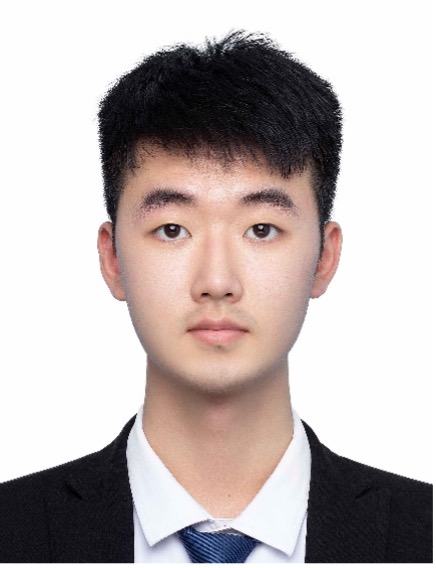}}]
{Sizhe Xing}
received his B.E. degree in school of optical and electronic information, Huazhong University of Science and Technology, Wuhan, China. He is currently working toward his Ph.D. degree at the School of Information Science and Technology, Fudan University.  His research interests include optical communication, integrated photonics, and free space communication.
\end{IEEEbiography}
\vspace{-10mm}

\begin{IEEEbiography}[{\includegraphics[width=1in,height=1.25in,clip,keepaspectratio]{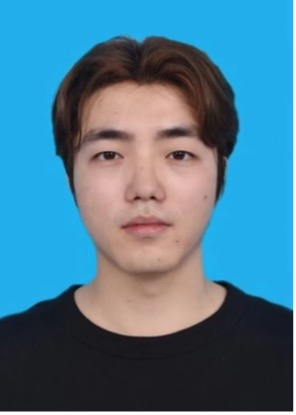}}]
{Yifan Sun}
is currently pursuing his Master's degree at the School of Information Science and Technology, Fudan University, Shanghai, 200438, China. He received his B.E. degree in school of Information Science and Technology, Fudan University, Shanghai, 200438, China. His research interests include optical communication and image transmission in multimode fiber.
\end{IEEEbiography}
\vspace{-10mm}

\begin{IEEEbiography}[{\includegraphics[width=1in,height=1.25in,clip,keepaspectratio]{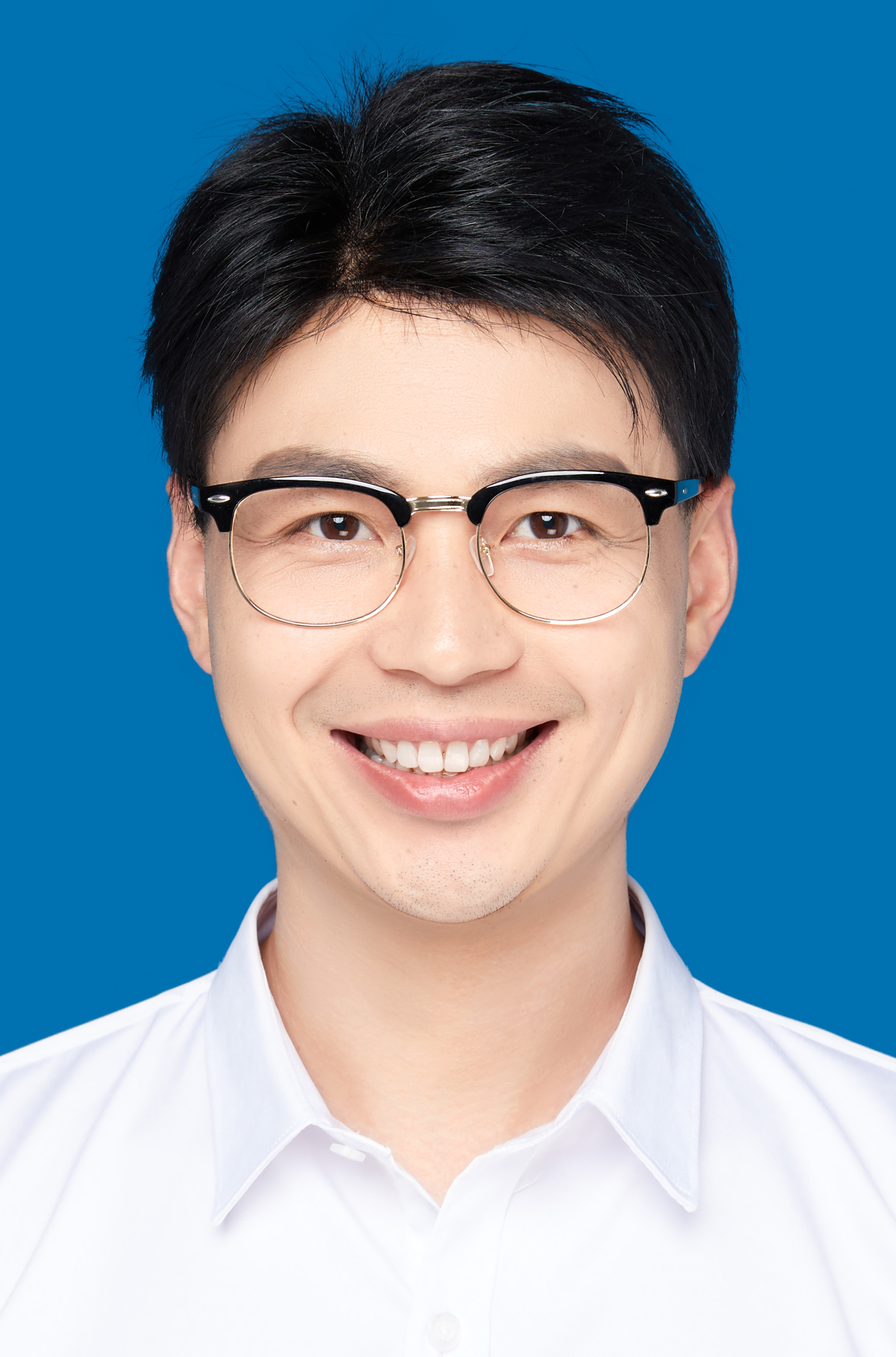}}]{Gangyong Jia} received the Ph.D. degree from the Department of Computer Science, University of Science and Technology of China, Hefei, China, in 2013. He is currently an Associate Professor with the Department of Computer Science, Hangzhou Dianzi University, China. His current research interests include the IoT, cloud computing, edge computing, power management, and operating
system.
\end{IEEEbiography}

\vfill

\end{document}